\documentclass[conference]{IEEEtran}
\IEEEoverridecommandlockouts
\usepackage{cite}
\usepackage{amsmath,amssymb,amsfonts}
\usepackage{algorithmic}
\usepackage{graphicx}
\usepackage{textcomp}
\usepackage{xcolor}
\def\BibTeX{{\rm B\kern-.05em{\sc i\kern-.025em b}\kern-.08em
    T\kern-.1667em\lower.7ex\hbox{E}\kern-.125emX}}

\usepackage{textcomp}
\usepackage{xcolor}
\usepackage{subcaption}
\usepackage{multirow}
\usepackage{booktabs}
\usepackage{cite}
\usepackage{algorithm}

\begin{document}

\title{Handwritten Text Segmentation  via End-to-End Learning of Convolutional Neural Networks\\
}

\author{\IEEEauthorblockN{1\textsuperscript{st} Junho Jo}
\IEEEauthorblockA{\textit{ Electrical and Computer Engineering} \\
\textit{INMC, Seoul National University}\\
Seoul, Korea \\
jottue@ispl.snu.ac.kr}
\and
\IEEEauthorblockN{2\textsuperscript{nd} Hyung Il Koo}
\IEEEauthorblockA{\textit{Electrical and Computer Engineering} \\
\textit{Ajou University}\\
Suwon, Korea \\
hikoo@ajou.ac.kr}
\and
\IEEEauthorblockN{3\textsuperscript{rd} Jae Woong Soh}
\IEEEauthorblockA{\textit{Electrical and Computer Engineering} \\
\textit{INMC, Seoul National University}\\
Seoul, Korea \\
soh90815@ispl.snu.ac.kr}
\and
\IEEEauthorblockN{4\textsuperscript{th} Nam Ik Cho}
\IEEEauthorblockA{\textit{Electrical and Computer Engineering} \\
	\textit{INMC, Seoul National University}\\
	Seoul, Korea \\
nicho@snu.ac.kr}
}

\maketitle

\begin{abstract}
We present a new handwritten text segmentation method by training a convolutional neural network (CNN) in an end-to-end manner. Many conventional methods addressed this problem by extracting connected components and then classifying them. However, this two-step approach has limitations when handwritten components and machine-printed parts are overlapping. 
Unlike conventional methods, we develop an end-to-end deep CNN for this problem, which does not need any preprocessing steps.
Since there is no publicly available dataset for this goal and pixel-wise annotations are time-consuming and costly, we also propose a data synthesis algorithm that generates realistic training samples. 
For training our network, we develop a cross-entropy based loss function that addresses the {\em imbalance} problems. 
Experimental results on synthetic and real images show the effectiveness of the proposed method.
Specifically, the proposed network has been trained solely on synthetic images, nevertheless the removal of handwritten text in real documents improves OCR performance from $71.13\%$ to $92.50\%$, showing the generalization performance of our network and synthesized images.
\end{abstract}


\section{Introduction}
\label{intro}

\begin{figure}[t]
	\begin{center}
		\begin{tabular}[t]{cc}
			\includegraphics[width=3.8cm]{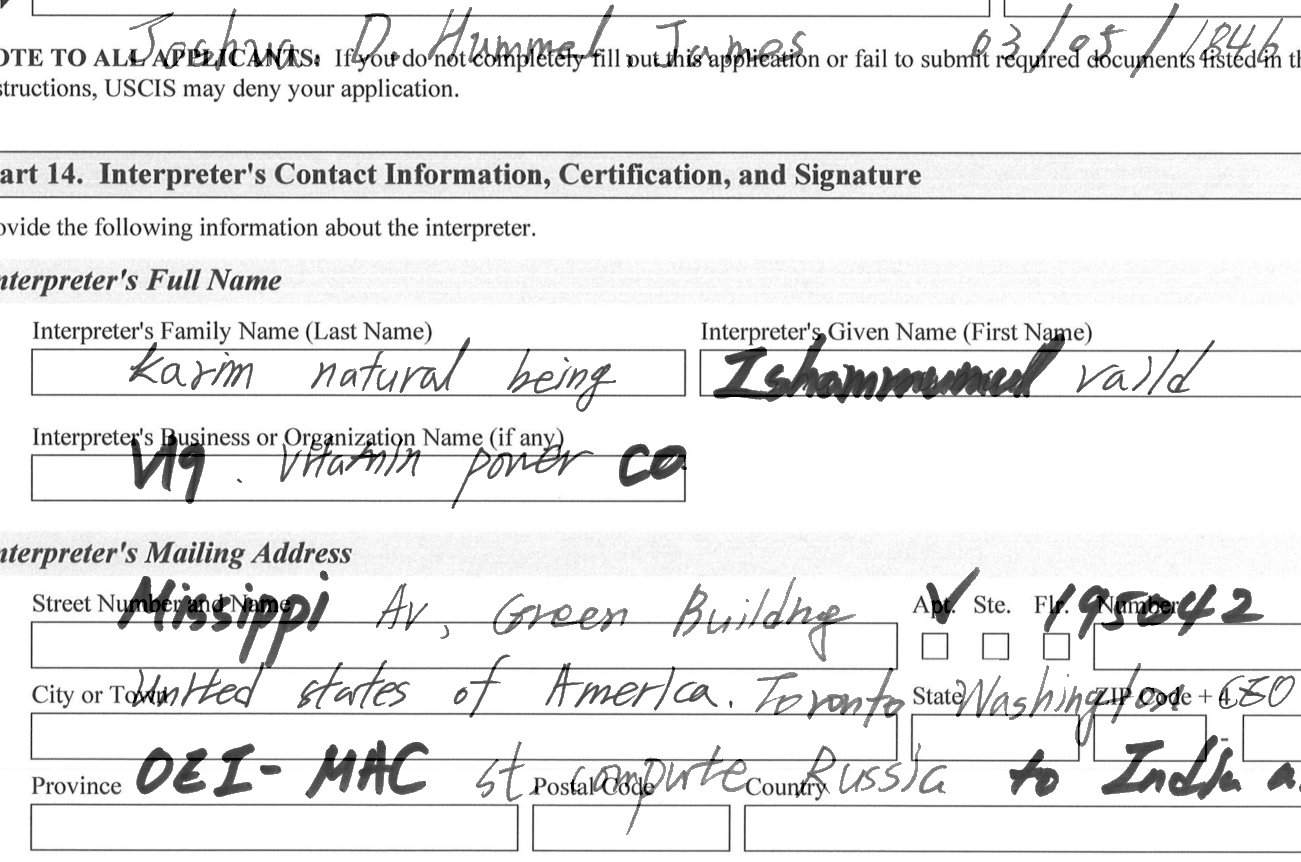}
			&\includegraphics[width=3.8cm]{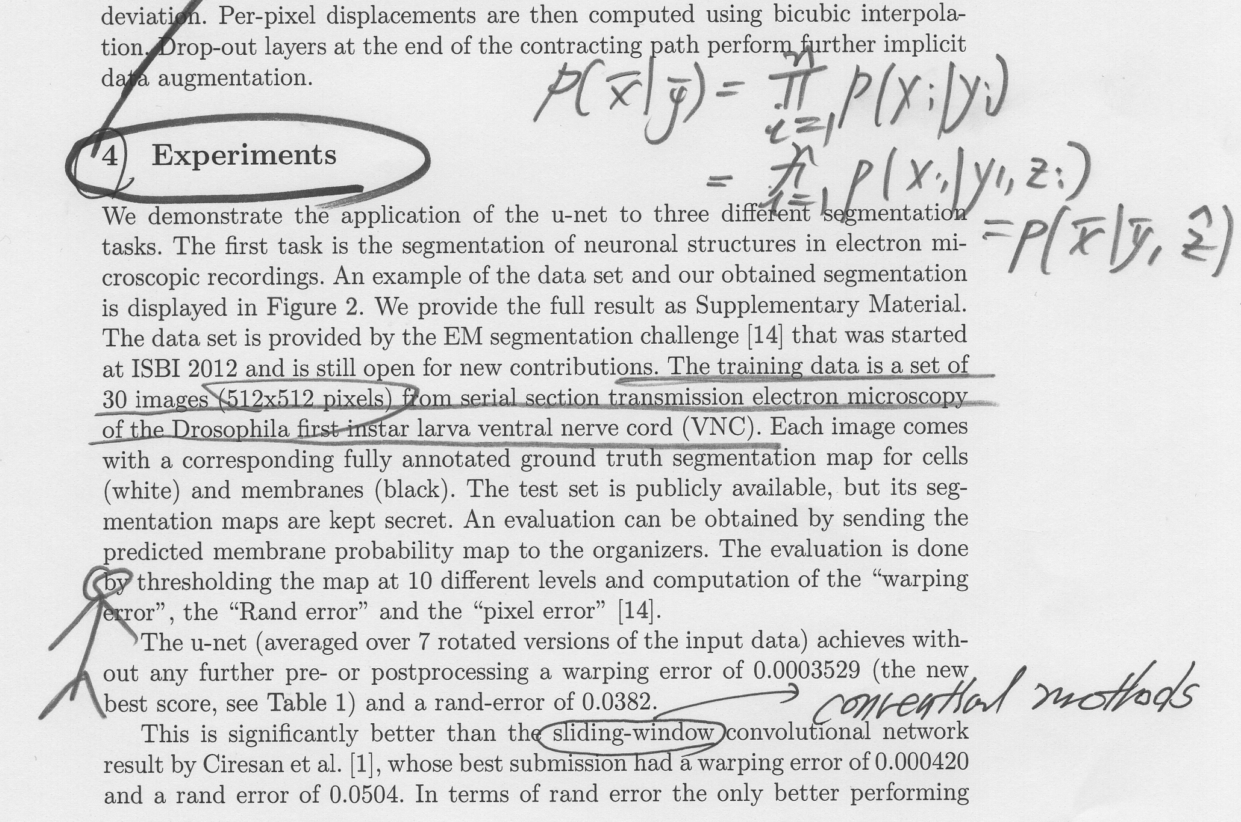} \\
			(a) & (b) \\
			\includegraphics[width=3.8cm]{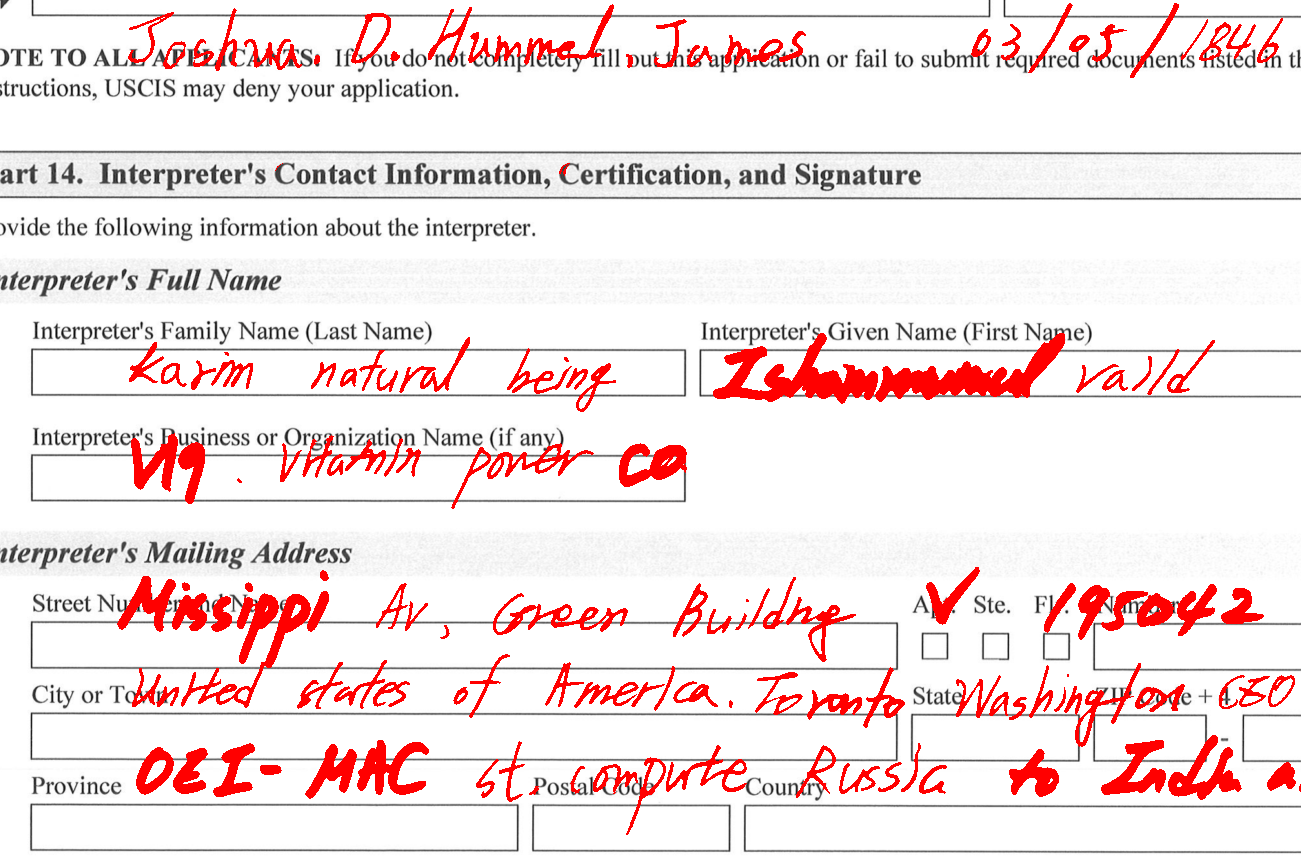} 
			&\includegraphics[width=3.8cm]{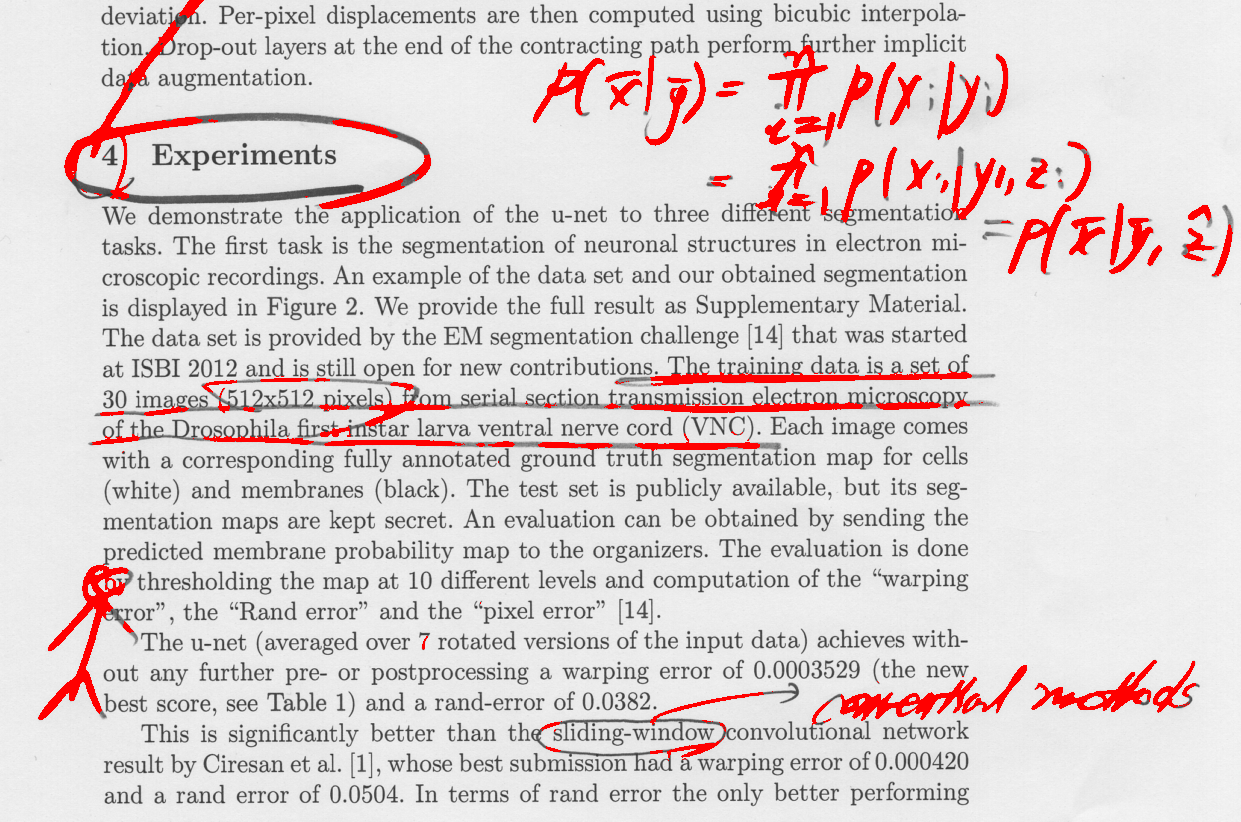} \\
			(c) & (d)
		\end{tabular}
	\end{center}
	\caption{The proposed method performs handwritten text segmentation from scanned document images: (a), (b) input images (synthetic and real), (c), (d) segmentation results (handwritten pixels are in red). %The proposed network was trained with synthetic but realistic images such as (a). 
	}
	\label{fig:example} % Fig. \ref{fig:example}
\end{figure}

Document digitization has been an important topic for the decades, and a huge number of methods have been proposed to address many kinds of sub-tasks such as optical character recognition (OCR), text-line segmentation, layout analysis, and so on \cite{b1,ww1,ww2}. 
Therefore, there have been many advances in machine-printed document understanding and handwritten text recognition.  However, the understanding of mixed cases ({\em i.e.}, documents having handwritten and machine-printed texts on the same page) still remains a challenging problem, especially when they are overlapping.
As shown in Fig.~\ref{fig:example}, these situations frequently occur in the formed documents, where we need to understand documents in the presence of handwritten notes and/or separate the handwritten and machine-printed texts.

\begin{figure*}[t]
	\centerline{\includegraphics[width=0.98\linewidth]{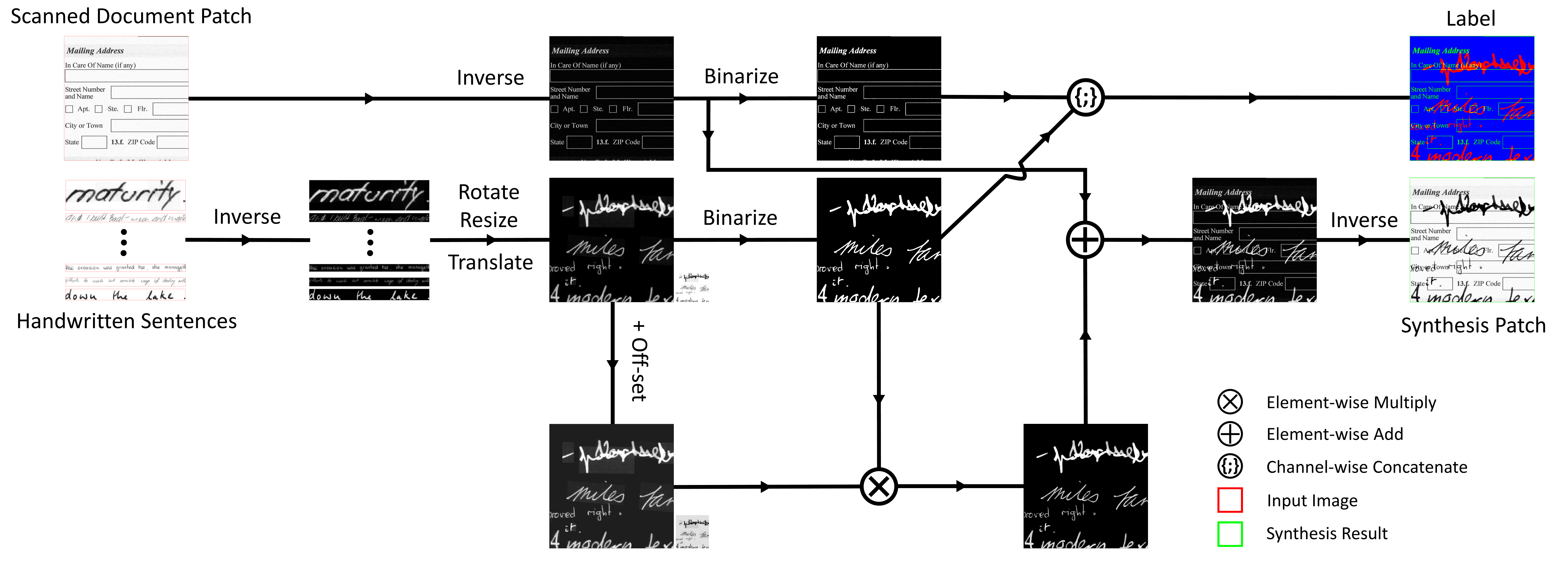}}
	\caption{Overall diagram of proposed data synthesis method.}
	\label{fig:synthesis_diagram}
\end{figure*}

Many researchers addressed this  problem  by separating  handwritten (or machine-printed) texts from the input  \cite{b2,b3,b4,b5,b6,b7,b8}. In \cite{b6}, they extracted connected components (CCs) and assigned feature vectors to them by exploiting the distribution of vectorized heights, widths, and distances between components. Finally, they classified each component by applying a $k$-nearest neighbor (NN) classifier.
Similarly, Kandan {\em et al.} \cite{b4} classified each component by using support vector machines (SVMs) and $k$-NN classifiers. Also, they improved descriptors so that the algorithm is robust to deformations. 
Recently, CNNs outperform traditional methods based on hand-crafted features in a variety of applications \cite{c1,c2,c3,c4}, and Li {\em et al.} used CNNs to classify  CCs \cite{b2}. Also, they incorporated conditional random fields into their framework to consider relations with neighboring CCs.

However, most of these conventional methods employed binarization and CC extraction as essential preprocessing steps and thus used the binary classification of CCs. These two-step approaches have advantages in that they allow us to exploit a variety of conventional modules (e.g., binarization, CC extraction, etc.), which also means that they have drawbacks that the final performance heavily depends on the performance of each module. Also, the CC extraction methods are prone to errors when two different kinds of texts are overlapping.
%, connected-components extracted from a binarized image can degrade the performance %since they are bundled as the same component.

To alleviate these problems, we propose a new handwritten text segmentation method based on an end-to-end CNN. To be precise, we formulate the task as a pixel-wise classification problem, assigning `$+1$' for pixels of handwritten text and `$-1$' for others (background, machine-printed text,  table boundaries, and so on).
For the segmentation network, we adopt the U-Net \cite{c3} that naturally exploits contextual information. 
In training the segmentation network, we address two challenges.
First, the number of handwritten text pixels is much smaller than the number of other pixels (mainly due to background pixels) \cite{e2}, so we develop a new loss function based on the conventional cross-entropy \cite{c1, c3}.
Second, since there is no publicly available dataset (the manual pixel-level annotation of documents is time-consuming and costly), we also propose a new synthesis method that can generate realistic images along with ground truth labels.

The  contributions of this paper can be summarized as follows:
\begin{itemize}
	\item  To the best of our knowledge, this is the first method that applies end-to-end learning to the separation of mixed-handwritten-machine-printed texts.
	\item For training the network under imbalanced situations, we propose a new loss function based on cross-entropy.
	\item For training, we also develop a data synthesis method, yielding realistic scanned documents as shown in Fig.~\ref{fig:synthesis_result}(b).
\end{itemize}

\section{Dataset synthesis} 
\label{dataset}

Although deep learning methods outperform conventional methods in many fields, training the deep networks requires a huge number of training samples. 
Especially, for the learning of segmentation networks (our application), pixel-level annotations are needed. 
However, there is no publicly available dataset for this goal, and we address this problem by synthesizing training samples

\subsection{Scan image dataset}

Synthesizing realistic images from scratch is a difficult task, and we develop a method that uses existing scanned images.
We use 13,353 sentence images of IAM dataset \cite{d1} as the handwritten texts, which was written by a variety of writers.  
For machine-printed parts, PRImA dataset \cite{d2} is used, which consists of scanned images of magazines and technical/scientific publications. 
Additionally, we manually crawled 141 images of questionnaire forms from the Internet. We augmented dataset by including these images in training dataset so that our dataset covers a wide range of formed documents. Typical examples are shown in $red$ boxes in  Fig.~\ref{fig:synthesis_diagram}.

\subsection{Dataset Synthesis}\label{syn}
\begin{figure}[t]
	\begin{center}
		\begin{tabular}[t]{cc}
			\includegraphics[width=4cm]{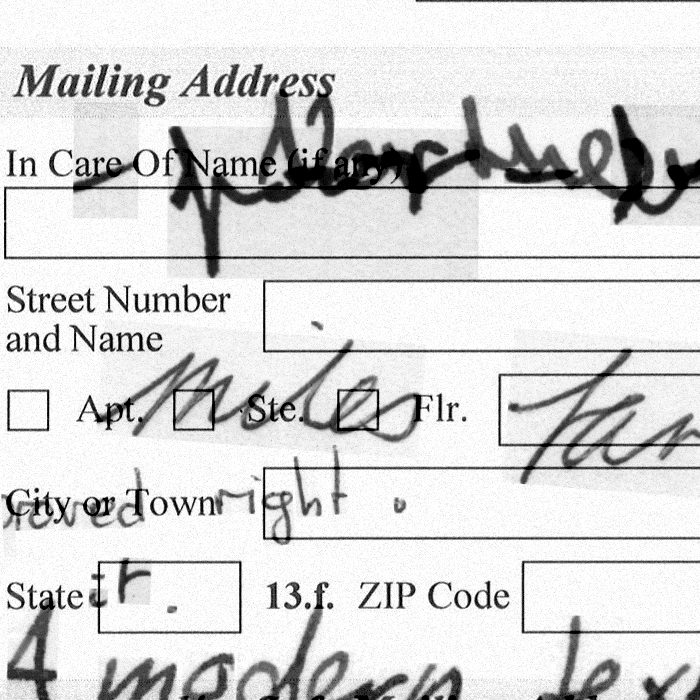}
			&\includegraphics[width=4cm]{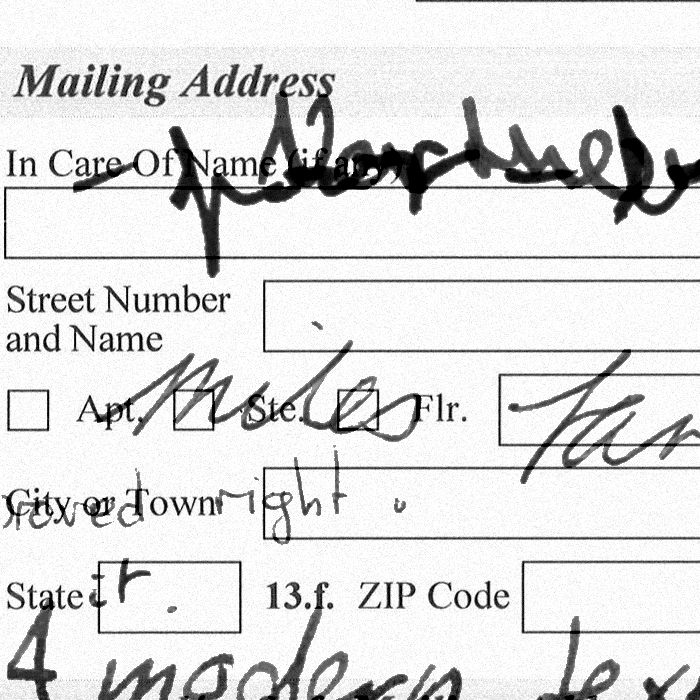} \\
			(a) & (b) 
		\end{tabular}
	\end{center}
	\caption{(a) synthesis result w/o artifacts handling, (b) synthesis result using the proposed method.}
	\label{fig:synthesis_result} % Fig. \ref{fig:example}
\end{figure}

For the realistic data synthesis, our main considerations are preserving textures of handwritten text images and noise of original documents. 
First, it is noted that textures of handwritten text images can be crucial evidence to differentiate them from the machine printed texts.
Secondly, consistent noise inherited from the scanning process must be preserved to diminish discrepancies between the distributions of synthetic and real data. To be precise, if we simply add a handwritten text image and a machine-printed text image, then the background will be saturated, and most of the scan-noises will disappear. We address this issue by inverting their intensities because backgrounds do not suffer from saturation if two inverted images are added. Another issue is undesirable block artifacts shown in Fig.~\ref{fig:synthesis_result}(a). Actually, these artifacts are from IAM dataset, since sentence images in IAM dataset were made by concatenating separate word images. 
To remove these artifacts, we extract only handwritten text pixels by multiplying images with its binary mask generated by {\em Otsu} binarization method \cite{otsu}. As shown in 
Fig.~\ref{fig:synthesis_result}(b), the proposed method yields realistic images, which have spatially consistent scan-noise.

In order to reflect the diversity of appearance observed in real environments, we apply randomized transformations, such as resizing, translation and rotation to each handwritten sentence image.
We also augment the dataset by adding random off-set values to the handwritten text patch to simulate a variety of intensities of handwritten texts. We have synthesized $146,391$ patches for training and $8,128$ for validation. The overall synthesis method is shown in Fig.~\ref{fig:synthesis_diagram}. 
We will make our synthesized dataset publicly available.

\section{Proposed method}
With synthesized training samples, we train a network in an end-to-end manner. This section presents our network structure and loss functions that are appropriately designed for our purpose and environment. 

\subsection{Network Structure} \label{section3}

\begin{figure}[t]
	\centerline{\includegraphics[width=0.98\linewidth]{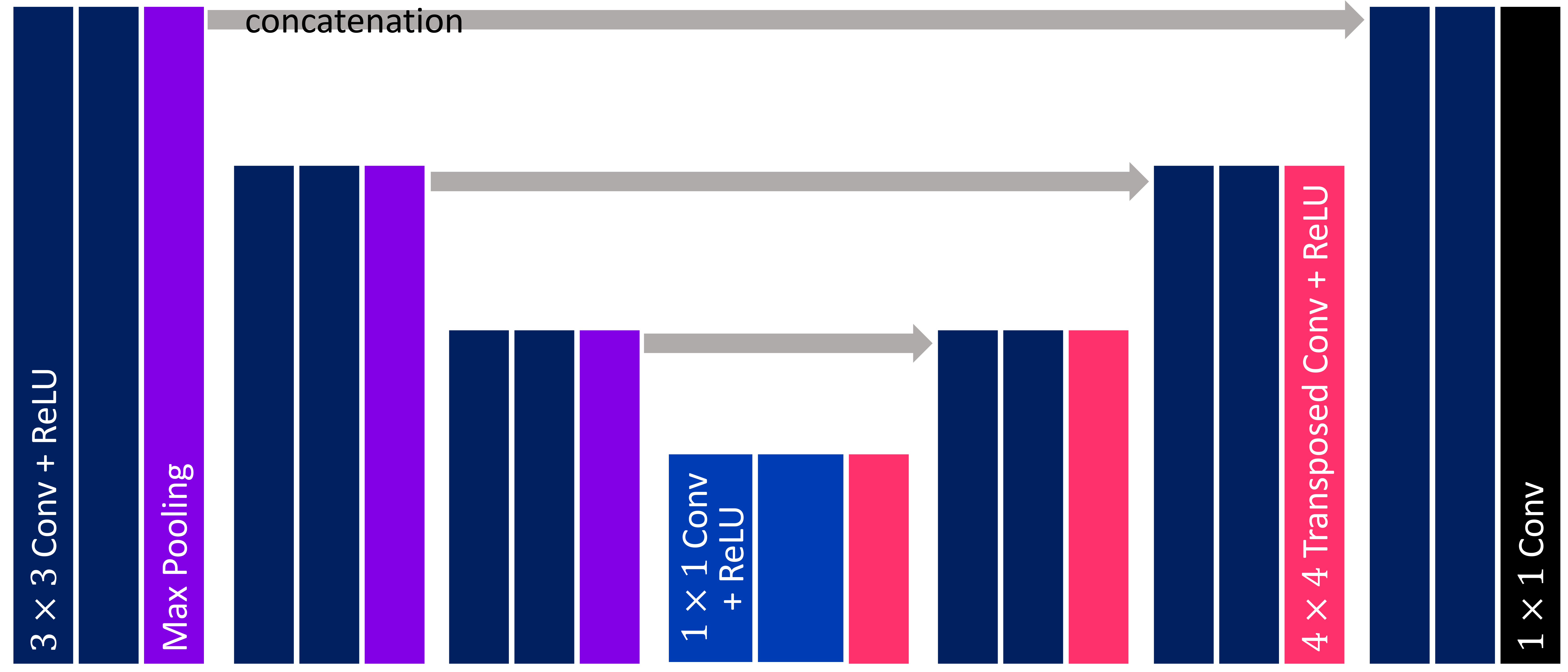}}
	\caption{Our network architecture based on U-Net.}
	\label{network}
\end{figure}

As a neural network architecture, we adopt U-Net \cite{c3} that consists of  encoder and decoder parts as shown in Fig.~\ref{network}.
The encoder captures the context of images by using a large-scale receptive field (downsampling operators), and the decoder generates high-resolution segmentation results by using contextual information and features from the encoder.
As downsampling operators, we empirically select max-pooling instead of strided-convolutions, and  $4\times4$ transposed-convolution layers with stride $2$ are used to up-sample the concatenation of encoder and decoder signals.

%%%%%%%%%%%%%%%%%%%%%%%%%%%%%%%%%%%%%%%%%%%%%%%%%%%%%%%%%%%%%%%%%%%%%%%%%%%%%%%%%%%%%%%%%%%%%%%%%%%%%%%%%%%%%%%%%%%%%%%%%%%%%%%%%%%%%%%%%%%%%%%%%%%%%%%%%%%%%%%%%%%%%%%%%%%%%%%%%%%%%%%%%%%%%%%%%%%%%%%%%%%%%%%%%%%%%%%%%%%%%%%%%%%%%%%%%%%%%%%%%%%%%%%%%%%%%%%%%%%%%%%%%%%%%%%%%%%%%%%%%%%%%%%%%%%%%%%%%%%%%%%%%%%%%%%%%%%%%%%%%
%As a neural network architecture, we adopt U-Net  that consists of  encoder  and  decoder parts \cite{c3}.
%As shown in Fig. \ref{network},
%the encoder captures the context of pixels 
%by using a large-scale receptive field (downsampling operators), and
%the decoder generates high-resolution segmentation results by using contextual information and features from the encoder. 
%
%{{\bf
%As the downsampling operators, we empirically select max-pooling instead of strided-convolutions. The decoder path consists of $3\times3$ convolution layers, $4\times4$ transposed-convolution layers with stride $2$ followed by concatenation with the corresponding feature maps from the encoder path, which enable precise localization.}}

%%%%%%%%%%%%%%%%%%%%%%%%%%%%%%%%%%%%%%%%%%%%%%%%%%%%%%%%%%%%%%%%%%%%%%%%%%%%%%%%%%%%%%%%%%%%%%%%%%%%%%%%%%%%%%%%%%%%%%%%%%%%%%%%%%%%%%%%%%%%%%%%%%%%%%%%%%%%%%%%%%%%%%%%%%%%%%%%%%%%%%%%%%%%%%%%%%%%%%%%%%%%%%%%%%%%%%%%%%%%%%%%%%%%%%%%%%%%%%%%%%%%%%%%%%%%%%%%%%%%%%%%%%%%%%%%%%%%%%%%%%%%%%%%%%%%%%%%%%%%%%%%%%%%%%%%%%%%%%%%%
\subsection{Cross entropy based loss function}\label{section4}
The cross-entropy loss function is commonly used for training the recognition \cite{c2,c4} and segmentation networks \cite{c1, c3}, which is described as 
\begin{equation}
\label{eq:CE2}
\mathcal{L}_{\mbox{\tiny CE}}(\theta) = \frac{1}{N}\sum_{n=1}^N \sum_{c=1}^C\mathbf{CE}(n,c),
\end{equation}
where 
\begin{equation}
\label{eq:CE1}
\mathbf{CE}(n,c) = -~t_{n,c}\log y_{n,c},
\end{equation}
where $n \in \left\lbrace 1,2,...,N \right\rbrace $ is the pixel index and $c \in \left\lbrace 1,2,...,C \right\rbrace $ is the class index. In our case, $C$ is set to $2$, since our model makes a decision just whether this pixel is handwritten text pixel or not.
Also, $t_{n,c}\in \left\lbrace 0,1 \right\rbrace $ and $y_{n,c}\in \left[0,1 \right] $ represent one-hot encoding of the ground truth label and the softmax result of the network, respectively. The $\theta$ denotes  parameters of the network. 

However, eq.(\ref{eq:CE2}) is likely to yield a poor local minimum for our task.  
That is, in most document images, the number of background pixels is approximately 20 times larger than that of text pixels. Therefore, the model is likely to converge to a trivial solution that classifies all pixels as background (the {\em class imbalance} problem).
Moreover,  background pixels consist of many easy cases and a tiny number of hard cases. In other words, most background pixels can be easily classified even by a simple thresholding method, and the CNN is very likely to converge to a sub-optimal solution by focusing on easy but dominant cases. That is, the loss summed over a large number of easy background examples would {\em overwhelm} the loss of rare hard examples, {\em i.e., ``many a little makes a mickle''} (the {\em overwhelming} problem).

%We will address these two problems in the following two %subsections respectively.  

%{\em overwhelming} problem is triggered by the fact that most of background %pixels are easily classified even by simple thresholding method, of course %more easily by CNN.
%Secondly, \emph{overwhelming} problem arises, since classification of the backgrounds is very easy task, even it is possible to classify using simple thresholding method.

\subsection{Dynamically Balanced Cross Entropy}
In order to alleviate the aforementioned {\em class imbalance} problem, we propose a new loss function:
\begin{equation}
\label{eq:DBCE}
\mathcal{L}_{\mbox{\tiny DBCE}}(\theta) = \frac{1}{N}\sum_{n=1}^{N}\sum_{c=1}^{C}\frac{1}{\beta(c) + \epsilon}~ \mathbf{CE}(n,c),
\end{equation}
where 
\begin{equation}
\label{eq:B}
\beta(c) = \frac{1}{N} \sum_{n=1}^{N}\left[t_{n,c}==1\right].
\end{equation}
%and
%\begin{equation}
% \sum_{c=1}^{C} \beta(c) = 1.
%\end{equation}
Unlike eq.(\ref{eq:CE2}), $\mathbf{CE}(n,c)$ is (dynamically) divided by the {\em frequency} of pixels in each class (in {\em mini-batch}) in $\mathcal{L}_{\mbox{\tiny DBCE}}(\theta)$. That is, the amount of contribution of each pixel is weighted by the scarcity of its class (fewer cases will have larger weights). By employing our loss function, the model can be trained even in imbalanced situations.
\begin{figure*}
	\begin{subfigure}{.245\textwidth}
		\centering
		\includegraphics[width=.98\linewidth]{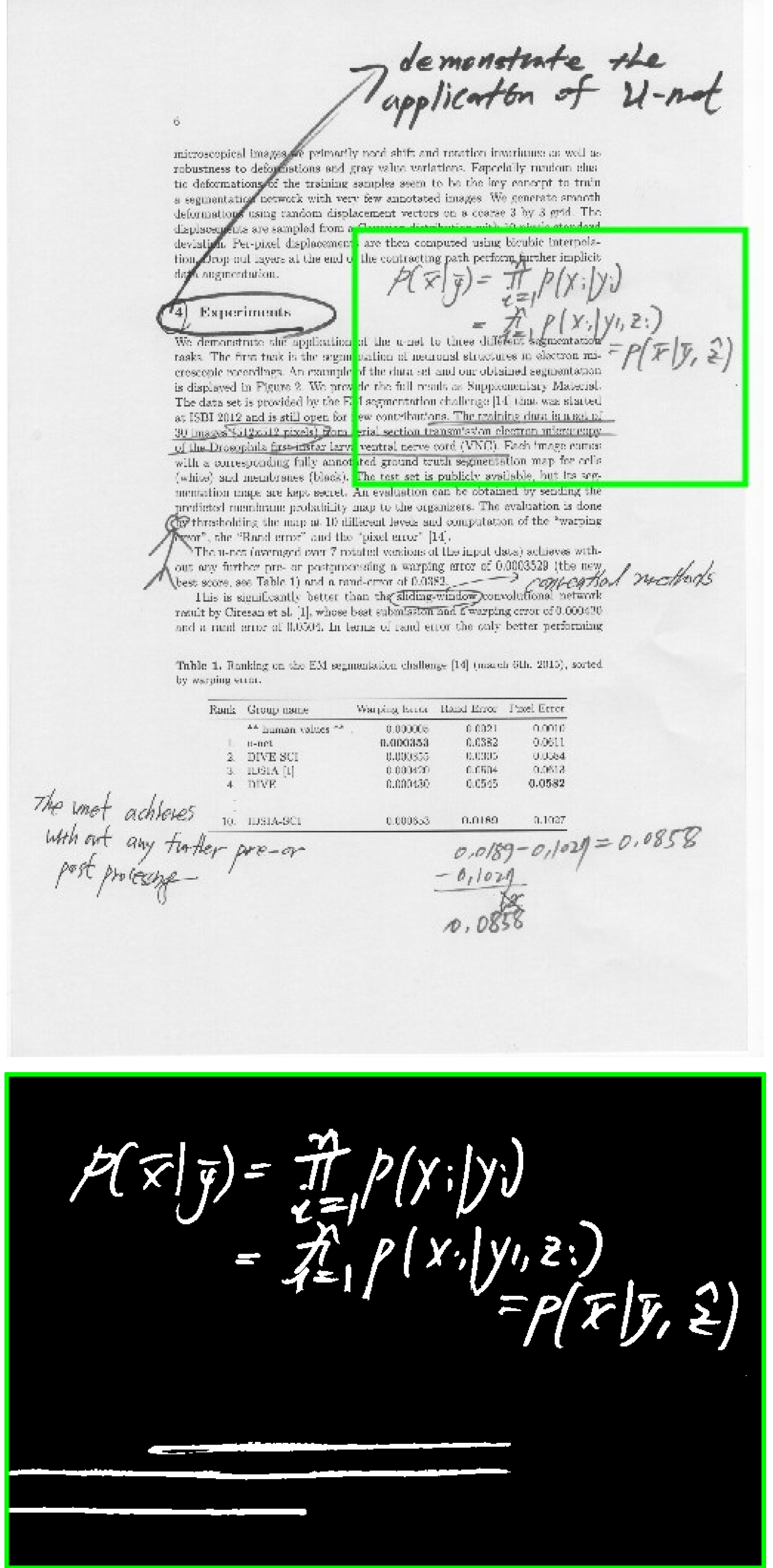}
		\caption{original image}
		\label{fig:sfig1}
	\end{subfigure}%
	\begin{subfigure}{.245\textwidth}
		\centering
		\includegraphics[width=.98\linewidth]{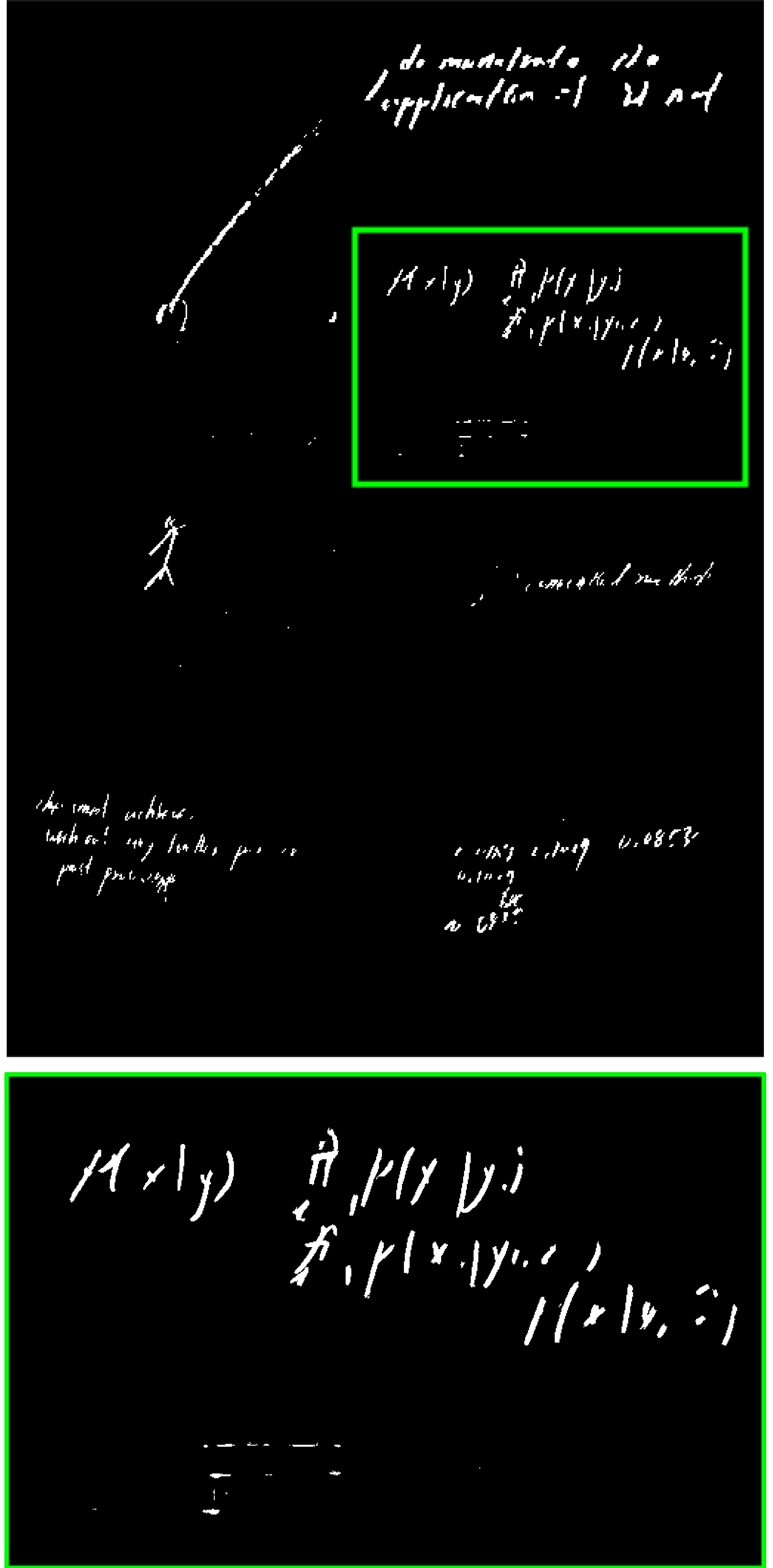}
		\caption{CE}
		\label{}
	\end{subfigure}
	\begin{subfigure}{.245\textwidth}
		\centering
		\includegraphics[width=.98\linewidth]{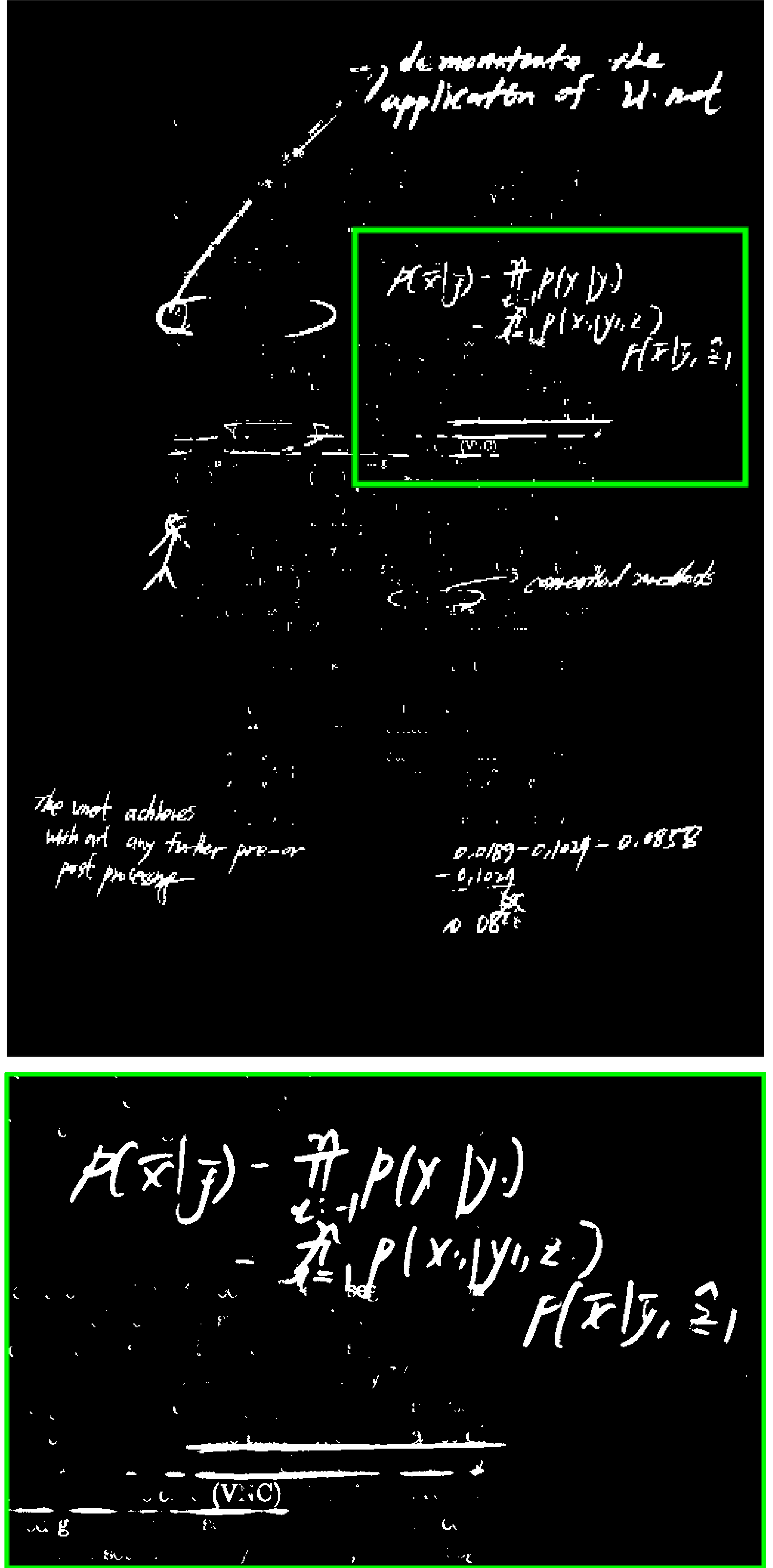}
		\caption{DBCE}
		\label{fig:sfig2}
	\end{subfigure}
	\begin{subfigure}{.245\textwidth}
		\centering
		\includegraphics[width=.98\linewidth]{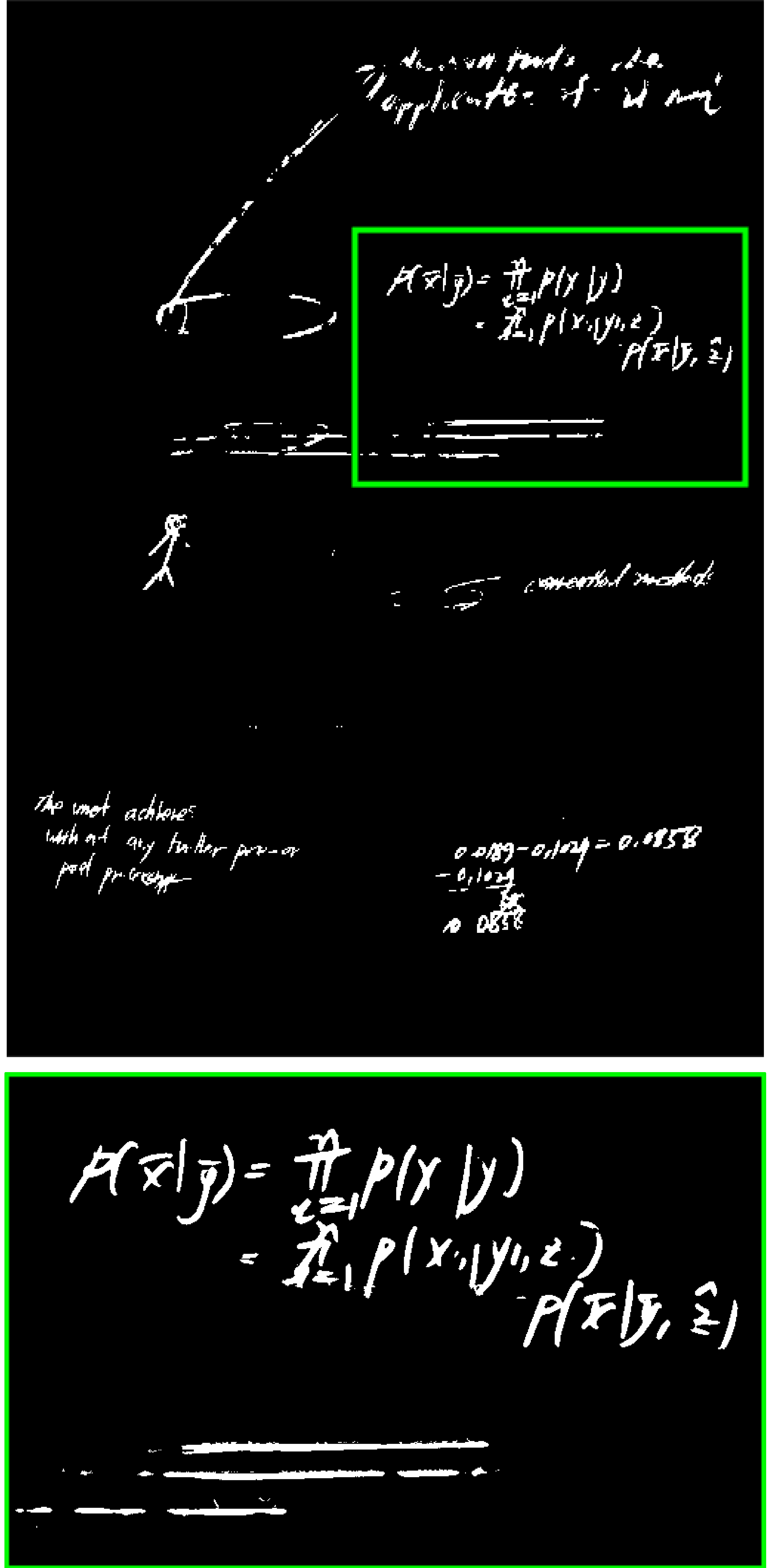}
		\caption{DBCE + F}
		\label{fig:sfig2}
	\end{subfigure}
	\caption{Segmentation results on a real scribbled document by the proposed network trained by each loss function. CE: conventional cross entropy, DBCE: dynamically balanced cross entropy, F: focal loss.}
	\label{fig:ablation_figure}
\end{figure*}

\subsection{Focal Loss}
In order to alleviate the {\em overwhelming} problem, we adopt focal loss in \cite{c4}. For the presentation of focal loss, we first define  $\mathbf{FCE}(n,c)$ as
\begin{equation}
\label{eq:FE}
\mathbf{FCE}(n,c) = -~({1-{ y}_{n,c}})^{\gamma}{ t}_{n,c} \log { y}_{n,c}, 
\end{equation}
where $\gamma$ is the hyperparameter that determines the boost degree of the penalty.
As shown in eq.(\ref{eq:FE}), the term $\mathbf{FCE}(n,c)$ is the $({1-{ y}_{n,c}})^{\gamma}$ scaled version of $\mathbf{CE}(n,c)$ in eq.(\ref{eq:CE1}).
This scaling factor automatically lessens the contribution of easy examples and makes the model focus on hard examples during the training. By putting two ideas together, the final loss function is given by
\begin{equation}
\label{eq:BFE}
\mathcal{L}_{\mbox{\tiny DBCEF}}(\theta)= \frac{1}{N}\sum_{n=1}^{N}\sum_{c=1}^{C}\frac{1}{\beta(c) + \epsilon} \mathbf{FCE}(n,c).
\end{equation}

\begin{figure}[t]
	\centerline{\includegraphics[width=0.98\linewidth]{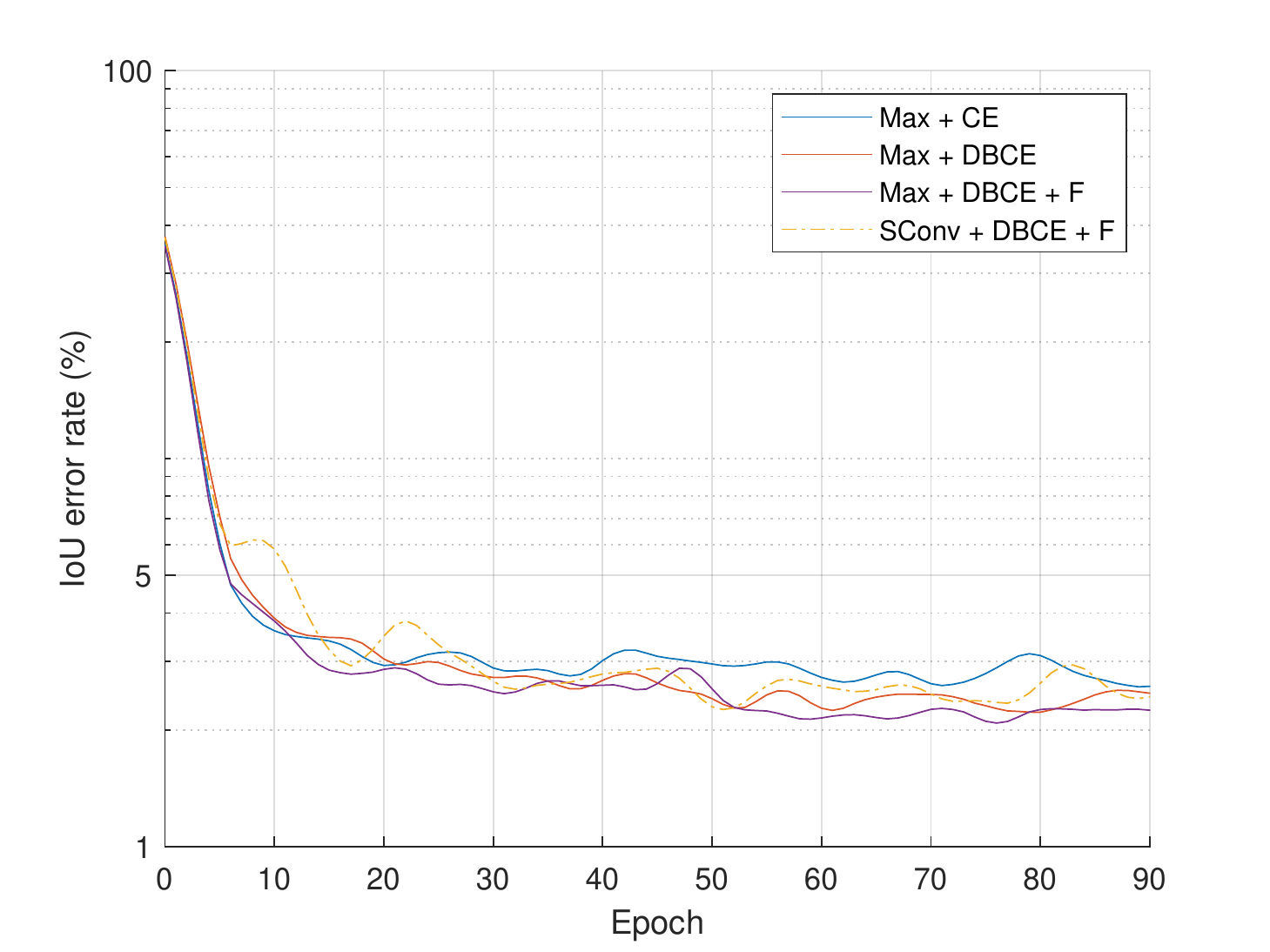}}
	\caption{IoU error rate of handwritten text in validation set. \textit{Max}: max-pooling,  \textit{SConv}: strided-convolution, \textit{CE}: conventional cross entropy, \textit{DBCE}: dynamically balanced cross entropy, \textit{F}: focal loss.}
	\label{fig:ablation}
	
\end{figure}

\subsection{Training details}
We have trained the network using Adam optimizer \cite{e1} with a mini-batch size of $32$. We used $0.0002$ as the initial learning rate with $0.8$ decay rate in every $30$ epochs. For hyper-parameter of {loss function}, we empirically set $\epsilon=0.0001$ and $\gamma= 1$.

%Effectiveness of our proposed cost function will be discussed in Section %\ref{sec:ablation}.
\begin{figure*}
	\begin{subfigure}{.24\textwidth}
		\centering
		\includegraphics[width=.98\linewidth]{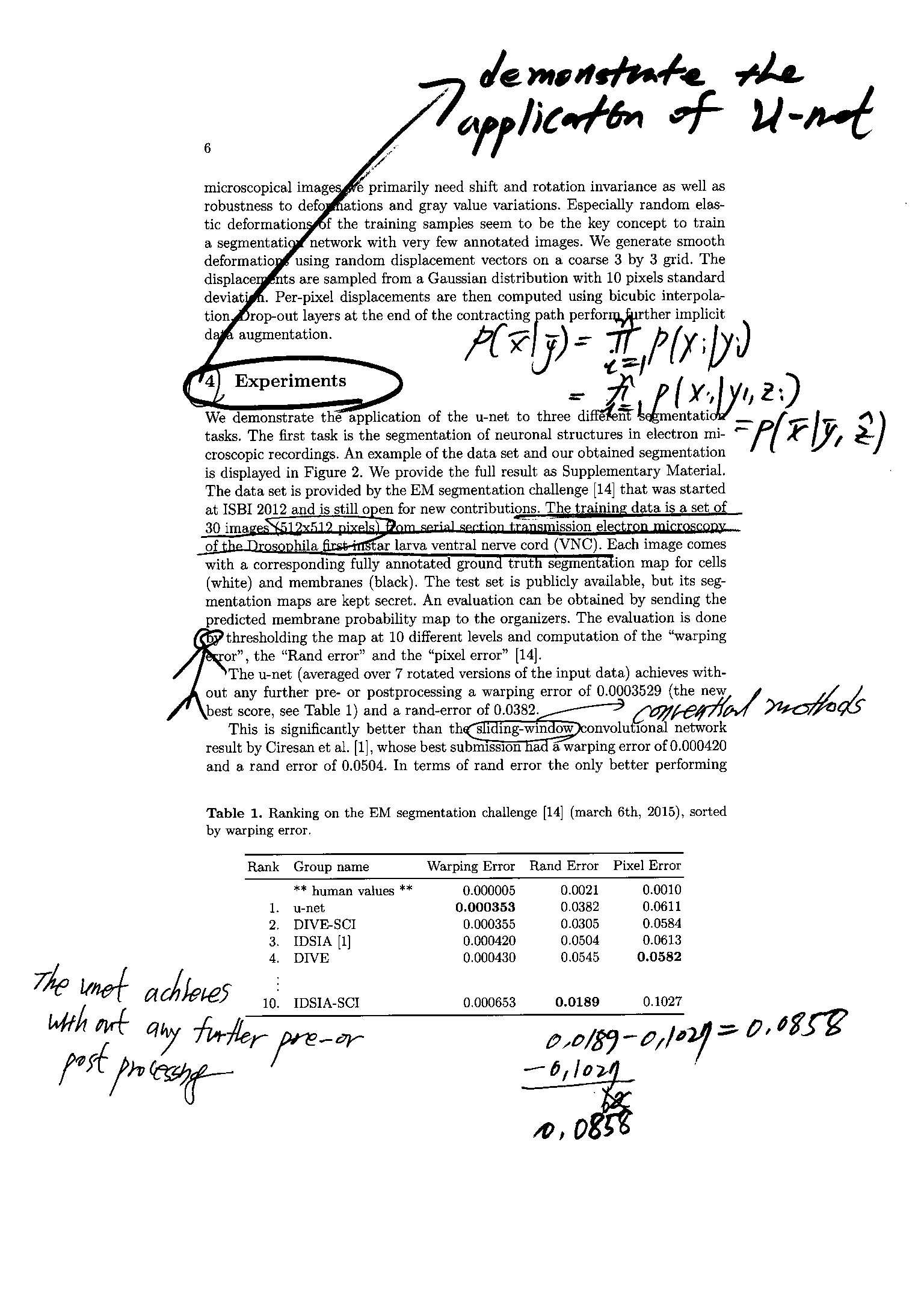}
		\caption{input 1}
		\label{fig:sfig1}
	\end{subfigure}%
	\begin{subfigure}{.24\textwidth}
		\centering
		\includegraphics[width=.98\linewidth]{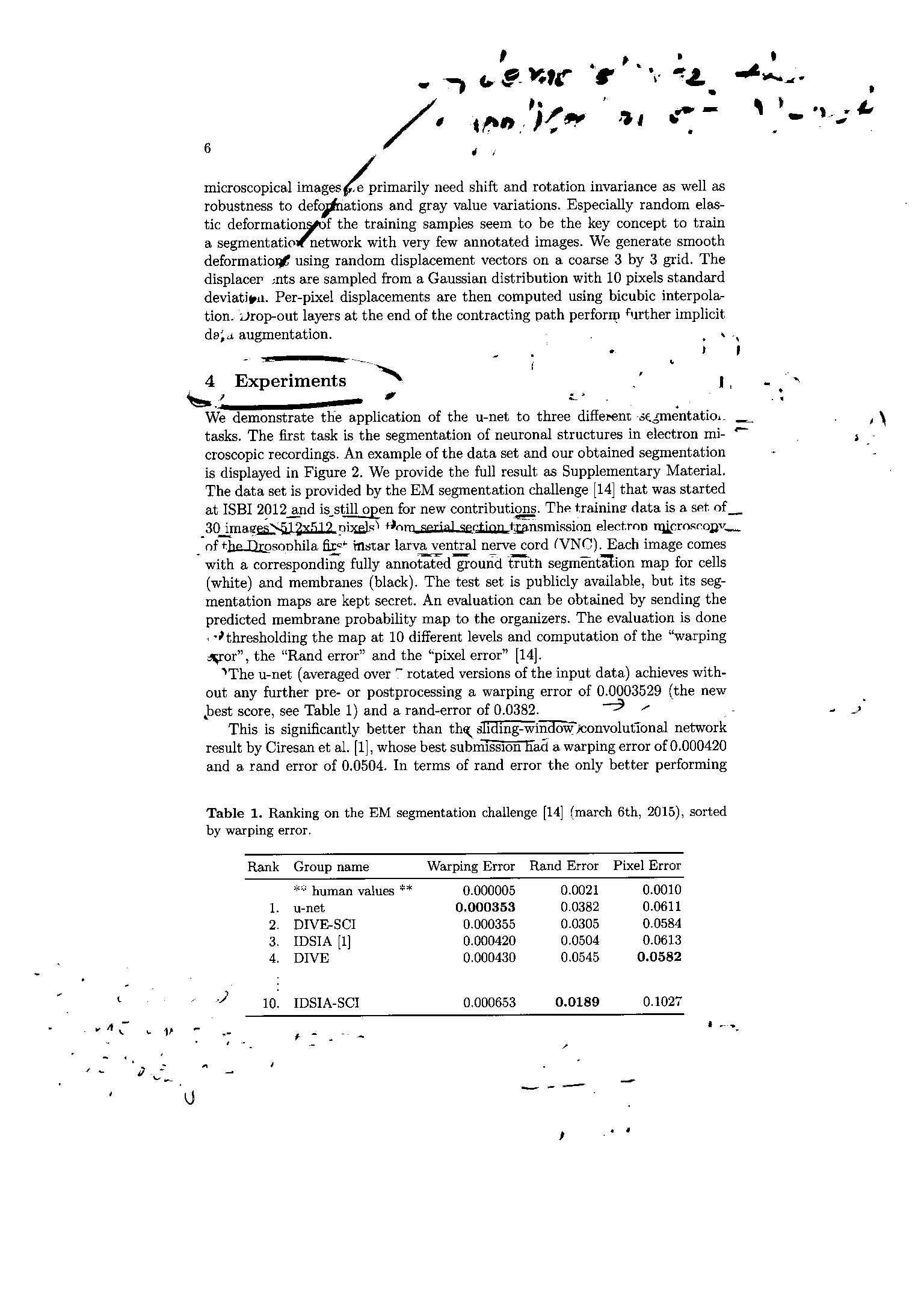}
		\caption{input 2}
		\label{}
	\end{subfigure}
	\begin{subfigure}{.24\textwidth}
		\centering
		\includegraphics[width=.98\linewidth]{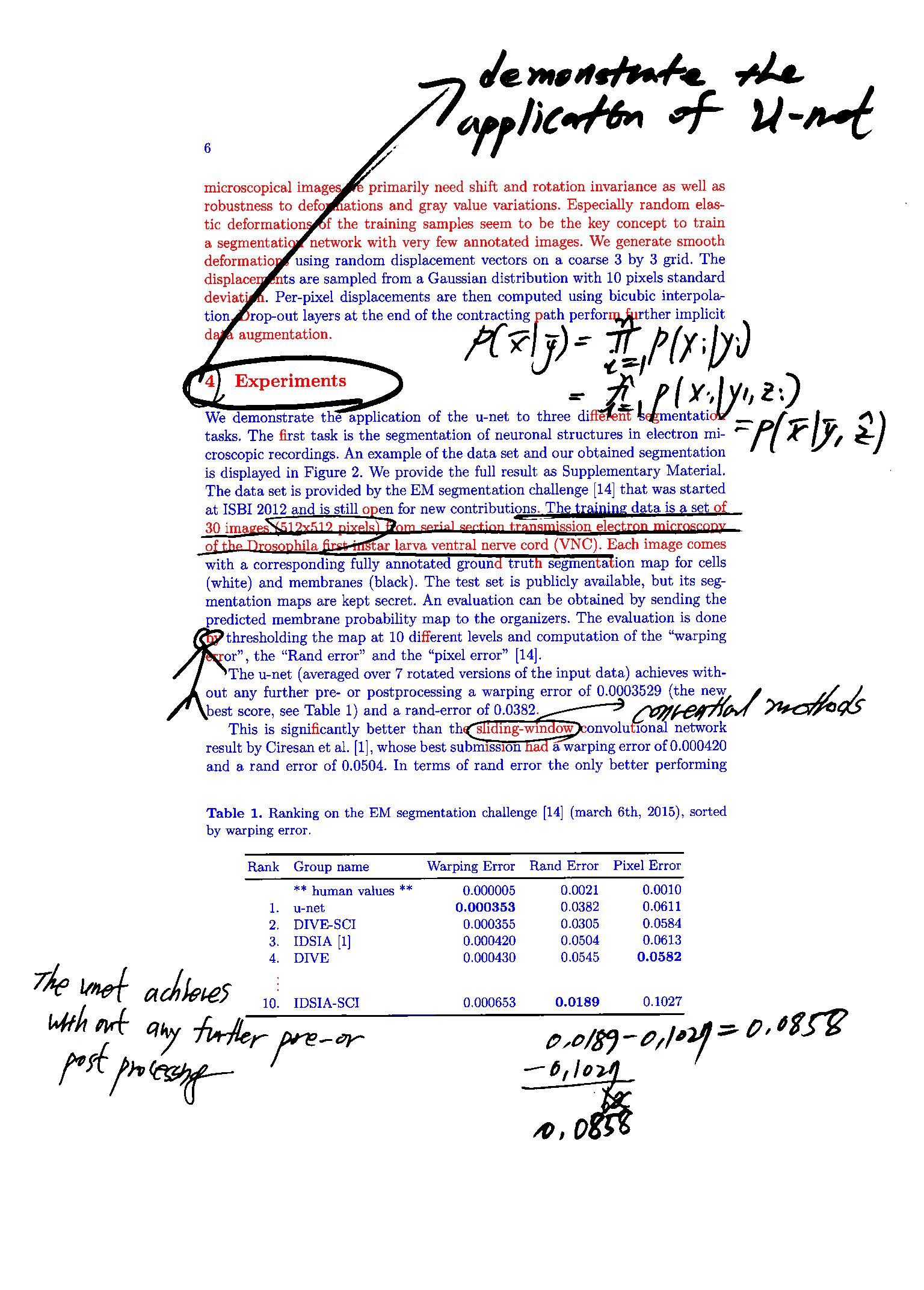}
		\caption{OCR result 1}
		\label{fig:sfig2}
	\end{subfigure}
	\begin{subfigure}{.24\textwidth}
		\centering
		\includegraphics[width=.98\linewidth]{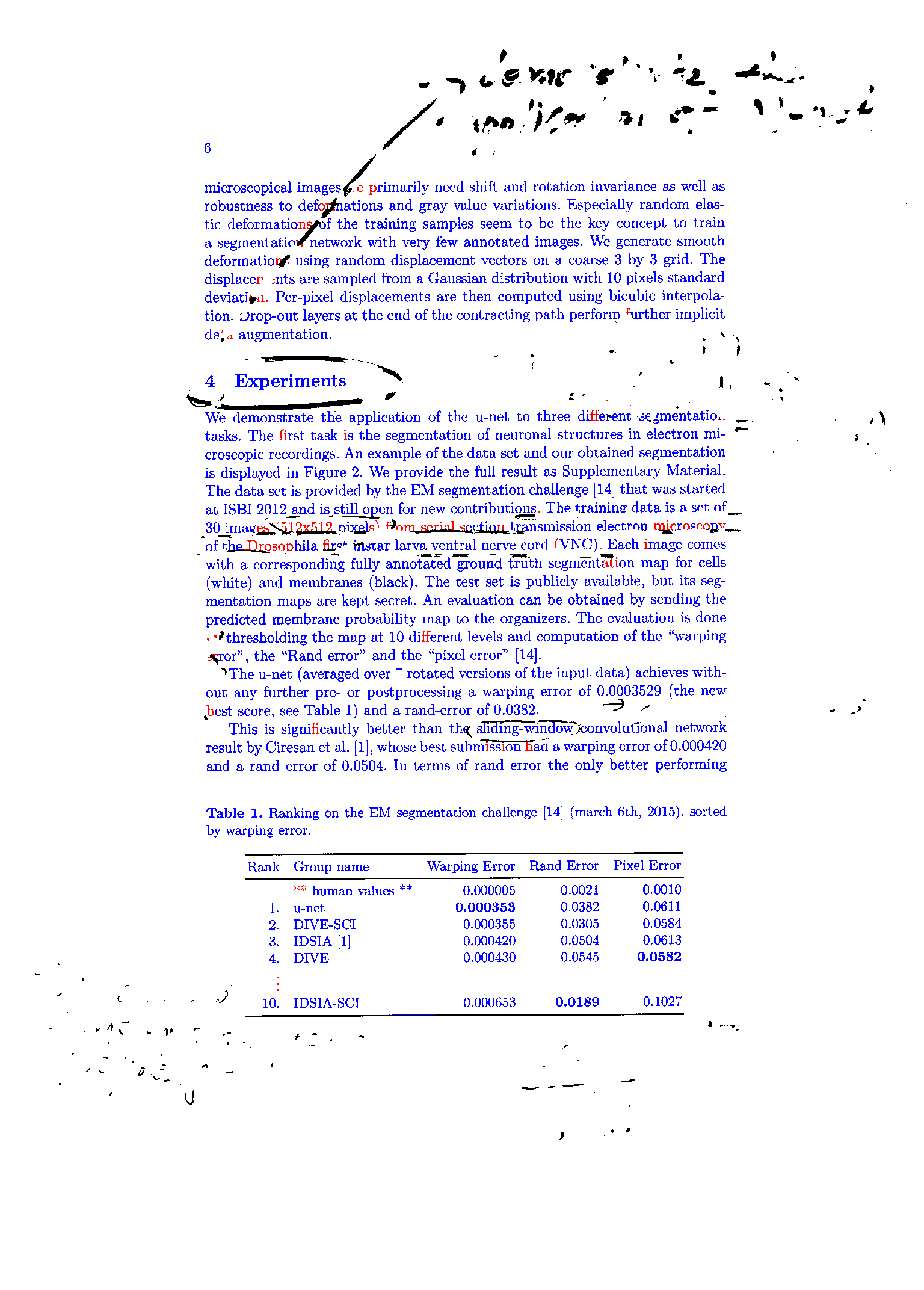}
		\caption{OCR result 2}
		\label{fig:sfig2}
	\end{subfigure}
	\caption{Comparisons of OCR performance. (a) Input 1 to OCR (real scribbled image), (b) Input 2 to OCR (removal of handwritten pixels from (a)), (c) and (d) OCR results for input 1 and 2 ({\color{blue}blue: correctly recognized characters}, {\color{red} red: missing or incorrect ones}).}
	\label{fig:OCR_result}
\end{figure*}

\section{Experimental Results} \label{section5}
In this section, we perform an ablation study to see whether our loss functions are effective for training the network. Then, we will show the performance of the proposed method on synthetic and real data. We did not perform the comparison with existing works \cite{b2,b3,b4,b5,b6,b7,b8}, since there is none that publicly provides the code and data to compare the performance. 
In a recent research of \cite{b2}, they tested CC-level and region-level segmentation with their own  {\em TestPaper 1.0} dataset, and the {\em Maurdor} dataset from \cite{m1}. However, these datasets are currently not accessible, and also they cannot be directly compared with ours because we deal with pixel-level results. Hence, we will instead make our dataset and codes publicly available for future research and comparisons.

\subsection{Ablation Study}
Fig.~\ref{fig:ablation_figure} demonstrates the effect of each loss function on the segmentation results, where the first column shows the input and the ground truth of segmentation results on green-box region. As shown in Fig.~\ref{fig:ablation_figure}(b), using the conventional cross entropy loss function ($\mathcal{L}_{\mbox{\tiny CE}}$) for training, a lot of handwritten pixels are classified as background, which means that the trained model experiences the {\em class imbalance} problem. 
By applying the proposed $\mathcal{L}_{\mbox{\tiny DBCE}}$, we can see that the {\em class imbalance} problem is quite mitigated as shown in Fig.~\ref{fig:ablation_figure}(c). However, there are lots of misclassified pixels in the machine printed text region, which is the example of {\em overwhelming} problem. 
To alleviate this problem, we incorporate focal loss \cite{c4} with proposed $\mathcal{L}_{\mbox{\tiny DBCE}}$. Finally, we can get the segmentation results without any degradation from {\em imbalance} situations as shown in Fig.~\ref{fig:ablation_figure}(d).

For the quantitative comparisons, we evaluate the proposed method using the pixel-level intersection-over-union (IoU) on synthesized sets. Fig.~\ref{fig:ablation} shows that the proposed loss function ($\mathcal{L}_{\mbox{\tiny DBCEF}}$)  improves validation performance.
Also, as shown in Table~{\ref{table}},  $\mathcal{L}_{\mbox{\tiny DBCE}}$ and focal loss term achieves $1.32\%$p and $0.60\%$p improvements of test performance on synthesized  set, respectively. These qualitative and quantitative experimental results showed that $\mathcal{L}_{\mbox{\tiny DBCE}}$ and focal loss term is meaningfully functioning during network training, {\em i.e.,} mitigating {\em class imbalance} and {\em overwhelming} problem well.

\begin{table}[t]
	\centering
	\caption{IoU results on synthesized test set. The best results are highlighted in \textbf{bold face} and the second best results are \underline{underlined}. \textit{H}: handwritten text, \textit{Max}: max-pooling, \textit{SConv}: strided-convolution, \textit{CE}: conventional cross entropy, \textit{DBCE}: dynamically balanced cross entropy, \textit{F}: focal loss}
	\label{table}
	\resizebox{0.9\linewidth}{!}{
		\begin{tabular}{lllclclcl}
			\toprule[1pt]
			\multicolumn{3}{c}{\textit{}} & \multicolumn{2}{c}{\multirow{3}{*}{\textit{\begin{tabular}[c]{@{}c@{}}Number of \\ Parameters \\ (M)\end{tabular}}}} & \multicolumn{4}{c}{\multirow{2}{*}{\textit{IoU (\%)}}} \\
			\multicolumn{3}{c}{\multirow{2}{*}{\textit{Model}}} & \multicolumn{2}{c}{} & \multicolumn{4}{c}{} \\ \cline{6-9}
			
			\multicolumn{3}{c}{} & \multicolumn{2}{c}{} & \multicolumn{2}{c}{\textit{non-H}} & \multicolumn{2}{c}{\textit{H}} \\ 
			\midrule[1pt]
			\multicolumn{3}{l}{\textit{Max + CE}} & \multicolumn{2}{c}{6.61} & \multicolumn{2}{c}{99.89} & \multicolumn{2}{c}{95.88} \\
			\multicolumn{3}{l}{\textit{Max + DBCE}} & \multicolumn{2}{c}{6.61} & \multicolumn{2}{c}{\underline {99.93}} & \multicolumn{2}{c}{\underline {97.20}} \\
			\multicolumn{3}{l}{\textit{\textbf{Max + DBCE +F}}} & \multicolumn{2}{c}{6.61} & \multicolumn{2}{c}{\textbf{99.94}} & \multicolumn{2}{c}{\textbf{97.80}} \\
			\multicolumn{3}{l}{\textit{SConv + DBCE + F}} & \multicolumn{2}{c}{7.39} & \multicolumn{2}{c}{99.92} & \multicolumn{2}{c}{97.11} \\ \hline
		\end{tabular}
	}
\end{table}

\begin{table}[t]
	\centering
	\caption{OCR performance on a real scribbled-document image. Accuracy is calculated by Correct / (Correct + Incorrect + Missing). Visualized results are shown in Fig.~\ref{fig:OCR_result}.}
	\label{table2}
	
	\begin{tabular}{clcllcllcll}
		\toprule[1pt]
		& \multicolumn{1}{c}{} & \multicolumn{9}{c}{Document states} \\ \cline{3-11} 
		& \multicolumn{1}{c}{} & \multicolumn{3}{c}{original} & \multicolumn{3}{c}{scribbled} & \multicolumn{3}{c}{separated} \\ \midrule[1pt]
		\multicolumn{2}{c}{Correct} & \multicolumn{3}{c}{2,141} & \multicolumn{3}{c}{1,584} & \multicolumn{3}{c}{2,071} \\
		\multicolumn{2}{c}{Incorrect} & \multicolumn{3}{c}{3} & \multicolumn{3}{c}{125} & \multicolumn{3}{c}{164} \\
		\multicolumn{2}{c}{Missing} & \multicolumn{3}{c}{7} & \multicolumn{3}{c}{518} & \multicolumn{3}{c}{4} \\
		\multicolumn{2}{c}{Accuracy (\%)} & \multicolumn{3}{c}{99.54} & \multicolumn{3}{c}{71.13} & \multicolumn{3}{c}{92.50} \\ \hline
		\multicolumn{1}{l}{} &  & \multicolumn{1}{l}{} &  &  & \multicolumn{1}{l}{} &  &  & \multicolumn{1}{l}{} &  &  \\
		\multicolumn{1}{l}{} &  & \multicolumn{1}{l}{} &  &  & \multicolumn{1}{l}{} &  &  & \multicolumn{1}{l}{} &  & 
	\end{tabular}
	
\end{table}

\subsection{Generalization performance}
%From a model point of view, images used to synthesized test set are never seen at all during training. Specifically, handwritten text images are made by ourselves. However, as shown in  \tablename~{\ref{table}}, the proposed model achieves good performance on the never-seen test set, $97.80\%$.

In the case of {\em real} test-set,  IoU evaluation is infeasible due to the lack of (pixel-level) ground truth. Rather, we measure the OCR performance on handwritten-pixel-removed-images, which is naturally proportional to the handwritten text segmentation performance.
To be precise, given a scribbled image like Fig.~\ref{fig:OCR_result}(a), we evaluate the OCR performances of original documents (w/o handwritten components) and handwritten-pixel-removed-images (Fig.~\ref{fig:OCR_result}(b)). 
As shown in Table~\ref{table2} and Fig.~\ref{fig:OCR_result}, there are a lot of missing or incorrectly detected characters in the scribbled document, mainly due to scribbles overlapping the machine-printed text. After removing them, which is segmented as handwritten text by proposed network, the OCR performance is improved from $71.13\%$ to $92.50\%$.
Note that the model is trained only with the synthesized data, and these results show that the model has learned features having generalization ability.

%%%%%%%
%We have evaluated the proposed method on synthesized and real sets.
%%For measuring the performance of our proposed network, we use two-kinds of images, which is {\em synthesized} images and {\em real scribbled} images.
%Synthesized test set consists  of  images generated by Algorithm \ref{algorithm1} and we have used the pixel-level intersection-over-union (IoU)  for the objective evaluation.
%In the case of {\em real} test-set,  IoU evaluation 
%was infeasible due to the lack of (pixel-level) ground truth. Rather, we measure OCR performance on handwritten-pixel-removed-images, which is proportional to the handwritten text segmentation performance.
%As shown in  \tablename~{\ref{table}}, the proposed model showed good performance on the test set, showing that the model has learned  features having generalization ability. Also, we evaluated OCR performance and the results are in
%\tablename~{\ref{table2}}. As shown, the OCR performance is improved from $73.71\%$ to $84.04\%$.
%
%
%
%In order to address the {\em class imbalance} problem, we replaced $\mathcal{L}_{\mbox{\tiny CE}}$ with $\mathcal{L}_{\mbox{\tiny DBCE}}$. As shown in \figurename~{\ref{ablation}} and \tablename~{\ref{table}}, the proposed $\mathcal{L}_{\mbox{\tiny DBCE}}$ improved the performance from $95.88\%$ to  $97.20\%$.	
%Also, we employed the idea of  focal loss   $\mathcal{L}_{\mbox{\tiny DBCE}}$ to address {\em overwhelming} problem. As shown in the \figurename~{\ref{ablation}}, focal loss improves the validation performance.
%%%%%%%%

\section{Conclusion}
In this paper, we have proposed a  method to separate handwritten text from the machine-printed documents based on a deep neural network. Unlike conventional methods, we proposed a method that works in an end-to-end manner, and addressed the {\em class imbalance} and {\em overwhelming} problems in the training phase.
The network was trained with synthetic training samples generated by the proposed synthesis method. 
The experimental results show that the proposed model also works well for real document images. Although the proposed method shows a good handwritten text extraction performance, it can  reconstruct only handwritten part when it is overlapping with the machine-printed text. As a future work, we will address the layer separation problem to reconstruct both components.

\end{document}